\documentclass[runningheads]{llncs}
\usepackage[T1]{fontenc}
\usepackage{graphicx}
\usepackage{amsmath,amssymb,amsfonts}
\usepackage{algorithm}
\usepackage{algorithmic}
\usepackage{multirow}
\usepackage{url}
\usepackage{array}
\usepackage{booktabs}
\title{Class-Conditioned Gaussian Mixture Modeling for Imbalanced Time Series Quantification%
\thanks{This is the author's accepted manuscript. The final authenticated 
version is available online at \texttt{https://doi.org/10.1007/978-981-92-1465-5\_44}. 
Published in R.\,C. Wong et al. (Eds.): PAKDD 2026, LNAI 16599, pp. 560--572, 2026.}}
\titlerunning{Class-Conditioned GMM for Imbalanced Time Series Quantification}

\author{
Md Shahriar Kabir\inst{1}\orcidID{0009-0004-4908-445X} \and
Mayesha Maliha R. Mithila\inst{1}\orcidID{0009-0002-0354-1702} \and
Anne H. H. Ngu\inst{1}\orcidID{0000-0002-5877-0230} \and
Myl\`ene C.\,Q. Farias\inst{1}\orcidID{0000-0002-1957-9943} \and
Byron Gao\inst{1}
}

\authorrunning{Kabir et al.}

\institute{Texas State University, United States\\
\email{cpi12@txstate.edu} (corresponding author)\\
\email{\{elx12,angu,mylene,bgao\}@txstate.edu}}

\providecommand{\orcidID}[1]{}

\begin{document}
\maketitle

\begin{abstract}
Quantification, estimating class prevalences in bags of unlabeled instances is vital in domains where aggregate statistics are more important than individual instance labels, such as biosignal monitoring, fall detection, and activity recognition. We investigate this issue in the challenging setting of imbalanced time series data and develop CC-GMNet-TS, a class-conditioned Gaussian mixture quantifier that combines a Transformer-based feature extractor with per-class latent mixtures. Unlike previous mixture-based quantifiers, which use a single Gaussian mixture shared by all classes, CC-GMNet-TS assigns each class its own compact mixture in a bounded latent space and scores segment embeddings against these class-specific components to create bag-level representations that emphasize rare but informative patterns. Bags are constructed from labeled pools using the Artificial Prevalence Protocol (APP) and prior shift bag sampling (PShift) to cover a wide range of class prevalence scenarios, and the model is trained end-to-end with a quantification-oriented loss. Experiments on three benchmarks: EMG Data for Gestures, SmartFallMM, and UCI-HAR show that CC-GMNet-TS achieves lower error across the three benchmarks compared to traditional aggregators and recent deep quantifiers, while ablations confirm the contributions of both the Transformer backbone and class-conditioned mixtures during PShift.
 
\end{abstract}

\section{Introduction}

Quantification, often known as prevalence estimation, is the challenge of forecasting the class distribution within a set (or \emph{bag}) of unlabeled instances. Unlike traditional classification, which assigns a label to each instance, quantification estimates how frequently each class occurs in aggregate. This is crucial in domains where decisions depend on prevalences rather than per-instance predictions and where detailed annotations are costly or unnecessary, such as disease monitoring, fall-risk assessment, or large-scale behaviour analysis. The problem becomes especially challenging for \emph{temporal}, \emph{imbalanced} data. Biomedical and human-activity time series (e.g., EMG, accelerometer, EEG) are multi-channel, non-stationary, and exhibit strong intra- and inter-subject variability. Rare but critical events (falls, pathological patterns) may account for less than 1\% of all observations, yet accurate estimation of their prevalence is essential. In addition, the class prior at deployment often differs from that seen during training (prior probability shift), degrading
methods that implicitly assume stationary priors.

Classical quantification methods such as Classify-and-Count (CC) and its variants---Adjusted CC (ACC), Probabilistic CC (PCC), and Probabilistic ACC
(PACC) treat quantification as a post-hoc correction of classifier outputs \cite{forman2005_cc,bella2010_acc}, while EM-based procedures adjust classifier scores to new priors by iteratively re-estimating prevalences from unlabeled samples \cite{saerens2002adjusting}. More recent \emph{deep quantification} methods learn bag-level prevalences end-to-end: RNN-based architectures such as QuaNet \cite{esuli2018_rnnquant} and the Deep Quantification Network (DQN) of Qi et al. \cite{qi2021_dqn} aggregate instance embeddings before regressing prevalences; histogram- and mixture-based methods such as HistNetQ and GMNet \cite{perez2025_histnetq,perez2025_gmnet} model score or feature distributions in latent space. In parallel, Transformer architectures \cite{vaswani2017attention,mithila2025ms} have demonstrated high performance on time series by capturing long range dependencies \cite{lim2021_tft}.

However, in many mixture-based deep quantifiers, deep quantifiers are largely class agnostic, where a single Gaussian mixture is shared across all classes, effectively modeling the marginal latent distribution \(p(z)\) rather than the class-conditional \(p(z \mid y=c)\). In imbalanced time series, this encourages mixture components to concentrate on majority-class modes, so minority and majority patterns are absorbed by the same clusters. As a result, changes in the prevalence of rare classes under prior probability shift barely affect the global mixture statistics, leading to biased prevalence estimates and low sensitivity to rare events. By contrast, class-conditioned mixtures are explicitly designed to approximate each \(p(z \mid y=c)\) separately, reducing cross-class leakage in latent space and aligning the bag representation with the standard prior-shift model \(p(z) = \sum_c \pi_c\, p(z \mid y=c)\). This per-class modeling is the key conceptual advantage over class-agnostic mixtures, which only approximate the marginal \(p(z)\).

In this paper, we propose \emph{CC-GMNet-TS}, a time-series–aware quantification framework that combines a Transformer-based feature extractor with \emph{class-conditioned} Gaussian mixture modeling in latent space. Our key idea is to (i) use a Transformer encoder to obtain segment-level embeddings tailored to bag-level prevalence estimation, and (ii) map these embeddings into a bounded latent space $[0,1]^d$ where each class is modeled by its own Gaussian mixture. Concretely, our contributions are as follows:
\begin{itemize}
    \item We introduce a lightweight Transformer feature extraction module that handles variable length segments via padding and attention masks, uses a learnable class token, and applies attention pooling to focus on quantification relevant time steps.
    \item We design a class-conditioned Gaussian mixture module in a bounded latent space, where each class owns its own small set of mixture components, and bags are represented via aggregated likelihoods under these class-specific components.
    \item  We evaluate CC-GMNet-TS on three benchmarks (EMG, SmartFallMM, UCI-HAR) using the Artificial Prevalence Protocol (APP) and PShift, showing lower error than classical aggregators and recent deep quantifiers, especially on heavily imbalanced datasets.
\end{itemize}

\section{Related Work}
\label{sec:related}

\paragraph{Quantification and Prior Probability Shift}

Quantification can be described as supervised learning with prior probability shift. Class-conditional distributions stay stable, but class priors vary between training and deployment. Early work focused on post-hoc corrections of classifier outputs, such as Classify-and-Count (CC) and Adjusted CC (ACC) and their probabilistic variants PCC and PACC \cite{forman2005_cc,bella2010_acc}, or EM-based adjustment of posterior probabilities \cite{saerens2002adjusting}. Later studies analyzed the statistical properties and consistency of these estimators, and surveys highlight that most traditional methods treat quantification as a post-processing step over a base classifier, assuming i.i.d.\ instances and well-calibrated scores \cite{gonzalez2017_survey,esuli2022concise}. These assumptions are often violated in temporally correlated biomedical time series, where rare events are easily missed or miscalibrated.

\paragraph{Deep Neural Quantification Models}

To overcome the constraints of traditional aggregators, numerous deep learning algorithms are trained to forecast prevalences straight from sets or bags. Esuli et al. presented QuaNet, a recurrent neural network that accumulates instance level embeddings and regresses class prevalences for sentiment quantification \cite{esuli2018_rnnquant}. Deep Quantification Networks (DQN) \cite{qi2021_dqn} train generic deep encoders with a regression head to reduce quantification-specific losses like Relative Absolute Error, proving robustness against prior probability shift on textual datasets. Latent space modeling has emerged as a promising quantification approach. \emph{HistNetQ} \cite{perez2025_histnetq} and \emph{GMNet} \cite{perez2025_gmnet} employ differentiable histograms and Gaussian mixture models to encode classifier scores and latent features. While these approaches perform well on static and tabular data, the mixing components are shared across classes, and the encoders are not tailored for multi-channel time series signals with varying segment lengths.

\paragraph{Time series Modeling and Rare Event Estimation}

Transformers have lately exhibited strong capabilities in time series forecasting and classification, capturing long-range temporal dependencies\cite{vaswani2017attention}. Notable variants include Temporal Fusion Transformers (TFT) \cite{lim2021_tft}, Informer \cite{zhou2021informer}, and PatchTST \cite{nie2022time}, which supports variable-length inputs using masking and has been successfully applied to demand forecasting and sensor analysis. Recent diffusion-based models have also been explored for generating healthcare time-series signals, e.g., TransConv-DDPM~\cite{kabir2025transconv}. Deep learning approaches for imbalanced time series often favor classification over quantification, using class-balanced losses, focal losses \cite{lin2017focal} that adaptively re-weight loss contributions based on example difficulty, or data augmentation to resolve imbalance. However, classification-centric objectives are not immediately compatible with prevalence estimation, resulting in inaccurate estimates for rare classes. Transformer-based time series encoders with class-conditioned latent mixture models to estimate bag-level prevalence under prior probability shift remain largely underexplored.


\section{Methodology}
\label{sec:methodology}

\begin{figure}[t]
    \centering
    \includegraphics[width=\linewidth]{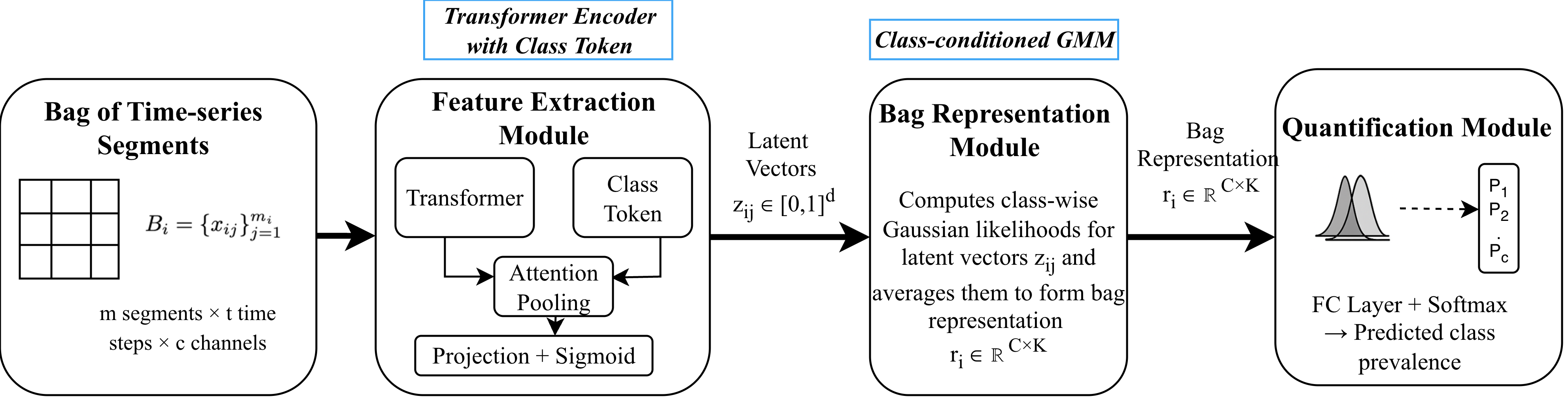}
    \caption{
    Overview of the proposed architecture. Bags of time series segments are encoded by a Transformer-based feature extraction module with a class token and attention pooling, summarized by a class-conditioned Gaussian mixture module into a bag representation, and finally mapped by a small fully connected head to bag-level class prevalences under weak supervision.
    }
    \label{fig:architecture}
\end{figure}

We address the problem of estimating class prevalences in imbalanced time series data using only bag-level supervision. Our framework is built around two main components: (i) a \emph{Transformer-based feature extraction module} that learns expressive latent representations from variable-length time series segments, and (ii) a \emph{class-conditioned Gaussian mixture module} that models the latent distribution of each class separately and aggregates instance-level features into a bag-level representation suitable for quantification. This design explicitly targets two challenges: handling segments with different lengths and improving sensitivity to rare classes under severe imbalance. An overview of the architecture is shown in Fig.~\ref{fig:architecture}.

\subsection{Problem Formulation}
\label{subsec:problem}

Let $\mathcal{Y} = \{1,\dots,C\}$ be the set of $C$ classes. Training data are provided as a collection of $N$ bags
\begin{equation}
\mathcal{D}_{\text{train}} = \big\{ (B_i, \mathbf{p}_i) \big\}_{i=1}^{N},
\end{equation}
where the $i$-th bag is a multiset of time series segments
\begin{equation}
B_i = \{ x_{ij} \}_{j=1}^{m_i}, \quad x_{ij} \in \mathbb{R}^{T_{ij} \times D}.
\end{equation}
Here, $T_{ij}$ is the number of time steps, $D$ is the number of input channels, and $m_i$ is the number of segments in bag $B_i$. The associated prevalence vector 
$\mathbf{p}_i = (p_{i1},\dots,p_{iC}) \in [0,1]^C$ satisfies $\sum_{c=1}^{C} p_{ic} = 1$ and encodes the proportion of each class in $B_i$. Instance-level labels are not used during training.

Given a new bag $B$, our goal is to learn a quantifier
\begin{equation}
f : B \longmapsto \hat{\mathbf{p}} \in [0,1]^C
\end{equation}
that estimates the true class prevalences in $B$ reliably, even when classes are highly imbalanced and the class prior at test time differs from that observed during training.

\subsection{Transformer-Based Feature Extraction Module}
\label{subsec:fem}

The first component of our framework is a Transformer-based feature extraction module (FEM) that maps each time series segment to a latent representation. This replaces recurrent encoders and is designed to better capture long-range temporal dependencies and to flexibly handle segments of different lengths.

Each segment $x_{ij} \in \mathbb{R}^{T_{ij} \times D}$ is first passed through a linear projection that maps each time step into a $d_{\text{model}}$-dimensional token:
\[
x_{ij}^{(t)} \in \mathbb{R}^{D}
\;\longrightarrow\;
\tilde{x}_{ij}^{(t)} \in \mathbb{R}^{d_{\text{model}}}, 
\quad t = 1,\dots,T_{ij}.
\]
We prepend a learnable class token to the sequence and add positional encodings to obtain the input to the Transformer encoder. For variable-length segments, we pad to a maximum length and use an attention mask so that padded positions do not influence the attention scores. In this way, the encoder can process bags that contain segments with different durations without changing the architecture.

The resulting sequence is processed by $L$ Transformer encoder layers, each consisting of multi-head self-attention followed by a position-wise feed-forward network, with residual connections and layer normalization. Self-attention allows the model to relate every time step to every other time step, which is particularly useful when discriminative patterns are long-range or occur at irregular positions in the segment.

To obtain a single vector per segment, we apply an attention-based pooling over the output tokens. Concretely, the Transformer produces a sequence of hidden states, and a small attention mechanism assigns a weight to each time step (including the class token). The segment embedding is then computed as a weighted sum of these hidden states, allowing the model to focus on the most informative parts of the time series for quantification. This combination of a class token with attention pooling is tailored to produce segment-level summaries that are directly useful for bag-level prevalence estimation, rather than just instance-wise classification.

Finally, we project the pooled representation into a $d$-dimensional latent vector and apply a sigmoid activation:
\begin{equation}
z_{ij} = \sigma\!\big( W_o \bar{h}_{ij} + b_o \big) \in [0,1]^d,
\end{equation}
where $\bar{h}_{ij}$ is the pooled Transformer output, and $W_o, b_o$ are learnable parameters. Bounding the latent features in $[0,1]^d$ improves numerical stability and simplifies the Gaussian modeling described next, since all components operate on a compact latent domain. It also regularizes feature scale, reducing outlier sensitivity in Gaussian likelihoods and stabilizing covariance learning.

\subsection{Class-Conditioned Bag Representation Module}
\label{subsec:brm}

The second key component is the class-conditioned bag representation module (BRM). Its goal is to aggregate the set of latent vectors $\{z_{ij}\}_{j=1}^{m_i}$ from bag $B_i$ into a fixed-dimensional representation that captures how likely the bag is under the latent distribution of each class. By modeling each class with its own mixture in latent space, the BRM explicitly encourages separated structure for majority and minority classes.

\paragraph{Class-specific Gaussian mixtures:}
We assume that, in the latent space, each class $c \in \{1,\dots,C\}$ is modeled by a Gaussian mixture with $K$ components. For class $c$, the $k$-th component has mean $\mu_{c,k} \in \mathbb{R}^{d}$ and covariance matrix $\Sigma_{c,k} \in \mathbb{R}^{d \times d}$. We use diagonal (or diagonal-plus-low-rank) covariances to keep the computation efficient and enforce positive definiteness by construction.

Given a latent vector $z_{ij}$ from bag $B_i$, we evaluate how likely it is under each component of each class using the standard Gaussian density
\[
p(z_{ij} \mid c,k) = \mathcal{N}\big(z_{ij} \mid \mu_{c,k}, \Sigma_{c,k}\big),
\quad c = 1,\dots,C,\; k = 1,\dots,K.
\]
Intuitively, these likelihoods measure how close $z_{ij}$ is to each class-specific cluster in the latent space, rather than mixing all classes into a single shared GMM.


\paragraph{Bag construction protocols:}
During training and evaluation, we construct bags from a pool of labeled
instances using two established sampling protocols.
\begin{itemize}
    \item \textbf{Artificial Prevalence Protocol (APP)}~\cite{esuli2022concise} samples a target class prevalence vector, converts it into counts per-class for a given bag size, and randomly draws that many instances from the labeled pool. Repeating this process yields bags that span uniform, skewed, and highly imbalanced class distributions.

    \item \textbf{Prior Shift Bag Sampling (PShift)}~\cite{gonzalez2024binary} follows the same steps but biases the sampled prevalence vectors toward more realistic or extreme mixtures, e.g., enforcing very rare or very frequent classes. This systematically shifts bag-level class priors while still using the same underlying pool.
\end{itemize}
Together, APP and PShift generate bags with diverse and controlled class prevalences, which we use consistently in both training and evaluation.

\paragraph{Bag-level representation:}
Given a bag $B_i$ constructed as above, we aggregate the likelihoods across all instances in $B_i$ to obtain a bag-level representation. For each class $c$ and component $k$, we average the likelihoods over $j$:
\[
r_{i,(c,k)} \;=\; \frac{1}{m_i} \sum_{j=1}^{m_i} p(z_{ij} \mid c,k),
\]
and collect all entries into a single vector
\[
r_i \in \mathbb{R}^{C \cdot K}.
\]
Thus, $r_i$ summarizes how strongly the instances in bag $B_i$ align with each Gaussian component of each class. This representation is permutation-invariant with respect to the order of segments inside the bag, and it explicitly encodes class-dependent structure in latent space. In particular, minority classes obtain their own mixture components, which reduces overlap with majority-class latent modes and improves robustness to rare but informative patterns.

\subsection{Quantification Module and Loss}
\label{subsec:qm}

The final quantification module takes the bag-level representation $r_i$ as input and outputs the predicted class prevalence vector $\hat{\mathbf{p}}_i$. We use a small multilayer perceptron:
\[
h_i = \phi(W_1 r_i + b_1), \qquad
u_i = W_2 h_i + b_2,
\]
where $\phi(\cdot)$ is a non-linear activation, and $W_1, b_1, W_2, b_2$ are learnable parameters. The final prevalence estimate is obtained with a softmax:
\begin{equation}
\hat{\mathbf{p}}_i = \text{Softmax}(u_i),
\end{equation}
which ensures $\sum_{c=1}^{C} \hat{p}_{ic} = 1$.

The model is trained end-to-end using a quantification-oriented loss that penalizes relative errors between predicted and true prevalences. In particular, we use the Mean Relative Absolute Error (MRAE),
\[
\mathcal{L}_{\text{MRAE}} 
= \frac{1}{N} \sum_{i=1}^{N} \frac{1}{C} \sum_{c=1}^{C}
\frac{|p_{ic} - \hat{p}_{ic}|}{p_{ic} + \epsilon},
\]
where $\epsilon > 0$ is a small constant to avoid division by zero and to control the influence of extremely rare classes. This loss directly optimizes the quality of prevalence estimates, rather than instance-wise classification accuracy.

\section{Experimental Results}

We evaluated our model on three imbalanced time series datasets: EMG \cite{uci_emg_dataset}, SmartFallMM \cite{smartfallmm_dataset}, and UCI-HAR \cite{human_activity_recognition_using_smartphones_240}. These datasets include the biosignal, fall detection, and activity recognition areas, with each presenting issues due to class imbalance and temporal relationships. We utilize MRAE \cite{gonzalez2017_survey} and NRAE \cite{sebastiani2020evaluation} as evaluation metrics, as well as Kullback-Leibler divergence (KLD) \cite{kullback1951information}. MRAE calculates the average relative deviation between predicted and true class prevalences, making it sensitive to deviations between minority classes. NRAE normalizes this value against a baseline (Classify-and-Count), allowing for consistent comparisons between datasets with varying class distributions. KLD measures the divergence between the predicted and true class distributions, as well as the information loss caused by approximation the true distribution.

\subsection{Datasets and Data Preparation}
\label{subsec:data}

In our methodology (Section~\ref{sec:methodology}), short time series windows are treated as segments $x_{ij}$ and then aggregated into bags $B_i$.
The EMG dataset includes recordings from 36 participants wearing a MYO wristband with 8 surface EMG sensors, executing 7 hand motions and a rest class (labeled 0--7). We discard timestamps and conduct global z-score normalization across all channels. Signals are segmented into 500-length windows with 50\% overlap. Window labels are determined via majority voting over frame-level labels, resulting in segments $x_{ij} \in \mathbb{R}^{500 \times 8}$. The rest class is dominant, resulting in a significant multiclass imbalance. SmartFallMM is a multi-modal human activity dataset gathered under IRB \#9461. We only used the 3D accelerometer stream (X, Y, Z) collected at 32 Hz from a smartwatch worn on the left wrist of 19 individuals. To build a binary task, map activity codes 10--14 to \emph{fall} and 1--9 to \emph{non-fall}. Signals are standardized per channel and split into non-overlapping windows of length 128, resulting in segments $x_{ij} \in \mathbb{R}^{128 \times 3}$ with significantly imbalanced fall and non-fall labels. The UCI-HAR dataset includes smartphone accelerometer and gyroscope readings from 30 people participating in six daily activities. We utilize the 128-length windows (2.56\,s at 50\,Hz) with 9 channels, standardize all channels, and adhere to the established 80/20 subject-wise train/test split. Each window is considered as a segment $x_{ij} \in \mathbb{R}^{128 \times 9}$ tagged with the activity class.

We create bags from these segments using the sample strategies described in Section~\ref{subsec:brm}. For the primary experiments, we create 200 training bags and 50 validation/test bags per dataset, each with 20 segments. In the Artificial Prevalence Protocol (APP)~\cite{esuli2022concise}, we sample a target prevalence vector (from a Dirichlet prior) and draw the corresponding number of segments per-class, resulting in bags with a diverse range of combinations. PShift~\cite{gonzalez2024binary} biases sampled prevalences towards extreme or skewed class proportions.  The Transformer encoder, the class-conditioned Gaussian mixture module, and the quantification head process all bags using $K=64$ Gaussian components per-class and latent dimension $d=256$, optimized with Adam (learning rate $10^{-4}$, weight decay $10^{-4}$), cosine annealing, and early stopping based on validation MRAE.

\subsection{Comparative Evaluation}

We compare CC-GMNet-TS against a wide range of quantification baselines, ranging from classical aggregators (CC, ACC, PCC)~\cite{forman2005_cc,bella2010_acc} to recent deep learning models: HistNetQ (histogram-based)~\cite{perez2025_histnetq}, DQN~\cite{qi2021_dqn}, QuaNet (RNN-based quantification network)~\cite{esuli2018_rnnquant} etc. All methods employ the same bag structure and training protocol. We employ the batch size=64, learning rate $1\mathrm{e}{-3}$ and weight decay $ 1\mathrm{e}{-4}$, using early stopping based on validation MRAE to select the best configuration for each method. Also, we used two Gaussian mixture layers, with 200 bags, each of size 20. Table~\ref{tab:mrae_nrae_kld} shows that CC-GMNet-TS achieves lower error than classical baselines and its class agnostic equivalent GMNet in terms of MRAE, NRAE, and KLD across all three datasets. The advantages are most obvious on EMG and SmartFallMM, where class-specific Gaussian components help to address severe class imbalance and complex temporal structure. On the more balanced UCI-HAR benchmark, CC-GMNet-TS achieves lower error than strong deep learning baselines such as TransNet and QuaNet across all measures, indicating that class-conditioned latent modeling can support reliable prevalence estimation across diverse time series scenarios.

\begin{table}[ht]
\centering
\footnotesize
\caption{Quantification performance across datasets. Lower is better for all metrics. Best metrics per dataset in \textbf{bold}.}
\label{tab:mrae_nrae_kld}
\begin{tabular}{l|ccc|ccc|ccc}
\toprule
\multirow{2}{*}{\textbf{Method}} &
\multicolumn{3}{c|}{\textbf{EMG}} &
\multicolumn{3}{c|}{\textbf{SmartFallMM}} &
\multicolumn{3}{c}{\textbf{UCI-HAR}} \\
\cmidrule(lr){2-4}\cmidrule(lr){5-7}\cmidrule(lr){8-10}
& MRAE & NRAE & KLD & MRAE & NRAE & KLD & MRAE & NRAE & KLD \\
\midrule
CC        & 0.530 & 0.300 & 0.200 & 0.520 & 0.260 & 0.195 & 0.490 & 0.280 & 0.208 \\
ACC       & 0.505 & 0.284 & 0.190 & 0.484 & 0.219 & 0.185 & 0.449 & 0.257 & 0.188 \\
PCC       & 0.460 & 0.260 & 0.180 & 0.430 & 0.210 & 0.175 & 0.400 & 0.240 & 0.178 \\
DQN         & 0.265 & 0.178 & 0.140 & 0.255 & 0.188 & 0.155 & 0.240 & 0.170 & 0.155 \\
QuaNet & 0.261 & 0.172 & 0.135 & 0.250 & 0.182 & 0.150 & 0.222 & 0.162 & 0.150 \\
TransNet  & 0.238 & 0.162 & 0.120 & 0.226 & 0.155 & 0.145 & 0.210 & 0.155 & 0.145 \\
MLQ     & 0.248 & 0.168 & 0.125 & 0.245 & 0.160 & 0.142 & 0.230 & 0.165 & 0.148 \\
HistNetQ & 0.280 & 0.188 & 0.150 & 0.313 & 0.195 & 0.165 & 0.278 & 0.179 & 0.165 \\
GMNet   & 0.240 & 0.162 & 0.138 & 0.225 & 0.153 & 0.140 & 0.223 & 0.160 & 0.142 \\
\textbf{CC-GMNet-TS}       & \textbf{0.232} & \textbf{0.158} & \textbf{0.115} & \textbf{0.211} & \textbf{0.146} & \textbf{0.135} & \textbf{0.204} & \textbf{0.151} & \textbf{0.132} \\
\bottomrule
\end{tabular}
\end{table}

\subsection{Ablation Study}

\paragraph{Effect of feature extractor and bag sampler:}
Table~\ref{tab:fe_bag_ablation} shows how the choice of feature extractor and bag sampler affects quantification accuracy. Across all three datasets, replacing the LSTM with a Transformer backbone reduces both MRAE and KLD, regardless of the sampling technique. On EMG, for example, MRAE drops from 0.242 to 0.223 when moving from \textit{LSTM+APP} to \textit{Transformer+APP}. Similarly, replacing APP with PShift generates consistent improvements for a fixed encoder. On all datasets, \textit{LSTM+PShift} improves over \textit{LSTM+APP}, and \textit{Transformer+PShift} improves over \textit{Transformer+APP}. These patterns are consistent with the components' inductive biases. The Transformer encoder can attend to all time steps and channels, capturing multi-scale temporal interactions that are difficult to retain in a single recurrent hidden state. This results in latent embeddings that are more separable for class-conditioned Gaussian mixtures. On the other hand, the PShift sampler actively generates training bags with more skewed and diversified class prevalences than APP, rather than focusing on moderate mixtures. This exposes the model to a broader range of imbalance scenarios that are more similar to prior changes seen during testing, resulting in more accurately calibrated prevalence estimations. The \textit{Transformer+PShift} combination consistently has the lowest MRAE and KLD on EMG, SmartFallMM, and UCI-HAR.

\begin{table}[t]
\centering
\footnotesize
\caption{Effect of feature extractor and bag sampling (MRAE $\downarrow$, KLD $\downarrow$)}
\label{tab:fe_bag_ablation}
\setlength{\tabcolsep}{1pt}
\begin{tabular}{lccccccc}
\toprule
Configuration & \multicolumn{3}{c}{MRAE} & \multicolumn{3}{c}{KLD} \\
\cmidrule(lr){2-4}\cmidrule(lr){5-7}
 & EMG & SmartFallMM & UCI-HAR & EMG & SmartFallMM & UCI-HAR \\
\midrule
LSTM+APP & 0.242 & 0.233 & 0.236 & 0.110 & 0.104 & 0.108 \\
LSTM+PShift & 0.231 & 0.225 & 0.229 & 0.097 & 0.093 & 0.095 \\
Transformer+APP & 0.223 & 0.218 & 0.224 & 0.089 & 0.086 & 0.090 \\
\textbf{Transformer+PShift} & \textbf{0.215} & \textbf{0.211} & \textbf{0.221} & \textbf{0.081} & \textbf{0.079} & \textbf{0.083} \\
\bottomrule
\end{tabular}
\end{table}

\begin{figure}[t]
    \centering
    \includegraphics[width=0.92\linewidth]{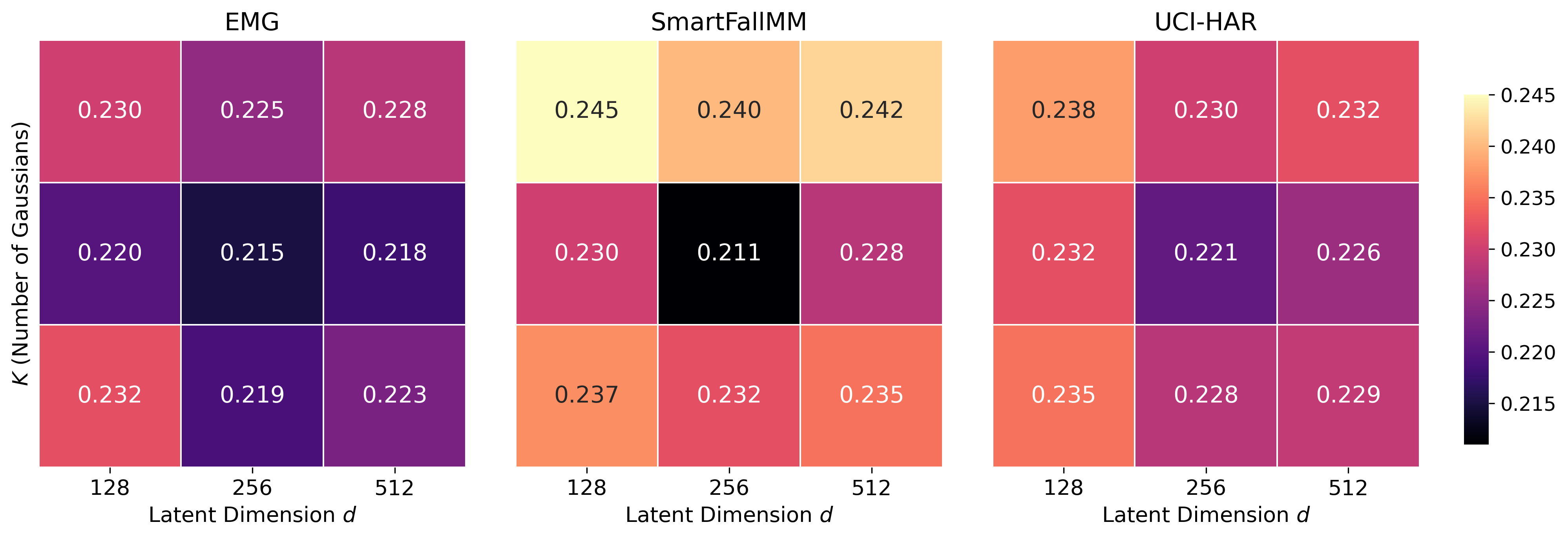}
    \caption{Effect of latent dimension $d$ and number of Gaussians per
    class $K$ on MRAE. Each heatmap shows the mean MRAE for one dataset;
    rows correspond to $K \in \{32,64,128\}$, columns to
    $d \in \{128,256,512\}$.}
    \label{fig:ablation_heatmap}
\end{figure}

\paragraph{Effect of latent dimension and number of Gaussians.}
We explore how the latent dimension $d$ and the number of Gaussians per-class $K$ effect performance by running all combinations of $K \in \{32, 64, 128\}$ and $d \in \{128, 256, 512\}$. The heatmaps in Figure~\ref{fig:ablation_heatmap} display MRAE for each dataset, with rows indexed by $K$ and columns by $d$. A similar trend appears from all three datasets. For \textbf{EMG}, the lowest error (0.215) occurs at $(K=64, d=256)$, whereas the remaining configurations lie in a narrow range between 0.218 and 0.232. SmartFallMM exhibits comparable behavior. $(K=64, d=256)$ again produces the lowest MRAE (0.211), with other values ranging from 0.228 to 0.245. For \textbf{UCI-HAR}, all combinations fall between 0.221 and 0.238, with the minimum attained at $(K=64, d=256)$. Intuitively, too few Gaussians or a low-dimensional latent space underfit intra-class variability, limiting how well class-conditioned mixtures can separate prevalence patterns, whereas too large $K$ or $d$ introduce redundant components and additional parameters that are difficult to estimate reliably from the available bags, resulting in noisier and less stable quantification. Overall, a moderate-capacity configuration of $K=64$ and $d=256$ provides the best consistent trade-off, and raising $K$ or $d$ does not result in significant gains.

\subsection{Discussion and Limitations}

We evaluate CC-GMNet-TS under \emph{prior probability shift} using APP/PShift-generated bags constructed from offline, pre-segmented windows. CC-GMNet-TS performs well across three different time-series domains (EMG gesture recognition, SmartFallMM fall detection, and UCI-HAR activity recognition), achieving lower MRAE and KLD than classical and recent deep baselines. Ablations provide two major design decisions: the Transformer backbone surpasses LSTM alternatives, and PShift is more effective than APP, particularly on imbalanced distributions. Hyperparameter study reveals that $K=64$ and $d=256$ offer a strong trade-off across datasets. However, there are certain restrictions that must be acknowledged. The model assumes a closed, fixed class set and operates offline on pre-segmented data, relying on synthetic bag generation procedures rather than realistic deployment distributions. Although these simplifications allow for controlled assessment, they do not take into account streaming, online, or uncertain label circumstances.

\section{Conclusion}

We presented CC-GMNet-TS, a class-conditioned Gaussian mixture framework for quantifying imbalanced time series data. The model directly enhances bag-level prevalence estimation by combining a Transformer-based encoder with per-class latent mixtures and a quantification-oriented loss. It also emphasizes unusual yet meaningful patterns. Experiments on three benchmarks (EMG, SmartFallMM, and UCI-HAR) reveal that CC-GMNet-TS achieves lower MRAE, NRAE, and KLD than classical aggregators and contemporary deep quantifiers, with notable improvements on highly imbalanced datasets. Ablations demonstrate the effectiveness of both the Transformer backbone and PShift. While our evaluation is limited to offline, pre-segmented recordings with closed class sets, the framework shows that class-conditioned latent modeling is a useful solution for temporal, highly imbalanced quantification.

\bibliographystyle{splncs04}
\bibliography{reference}

@inproceedings{forman2005_cc,
  title={Counting positives accurately despite inaccurate classification},
  author={Forman, George},
  booktitle={European conference on machine learning},
  pages={564--575},
  year={2005},
  organization={Springer}
}

@inproceedings{bella2010_acc,
  title={Quantification via probability estimators},
  author={Bella, Antonio and Ferri, Cesar and Hern{\'a}ndez-Orallo, Jos{\'e} and Ramirez-Quintana, Maria Jose},
  booktitle={2010 IEEE International conference on data mining},
  pages={737--742},
  year={2010},
  organization={IEEE}
}

@article{gonzalez2017_survey,
  title={A review on quantification learning},
  author={Gonz{\'a}lez, Pablo and Casta{\~n}o, Alberto and Chawla, Nitesh V and Coz, Juan Jos{\'e} Del},
  journal={ACM Computing Surveys (CSUR)},
  volume={50},
  number={5},
  pages={1--40},
  year={2017},
  publisher={ACM New York, NY, USA}
}

@inproceedings{qi2021_dqn,
  title={A framework for deep quantification learning},
  author={Qi, Lei and Khaleel, Mohammed and Tavanapong, Wallapak and Sukul, Adisak and Peterson, David},
  booktitle={Joint European Conference on Machine Learning and Knowledge Discovery in Databases},
  pages={232--248},
  year={2020},
  organization={Springer}
}

@article{perez2025_histnetq,
  title={Quantification using permutation-invariant networks based on histograms},
  author={P{\'e}rez-Mon, Olaya and Moreo, Alejandro and Coz, Juan Jos{\'e} del and Gonz{\'a}lez, Pablo},
  journal={Neural Computing and Applications},
  volume={37},
  number={5},
  pages={3505--3520},
  year={2025},
  publisher={Springer}
}

@article{perez2025_gmnet,
  title={Quantification via Gaussian Latent Space Representations},
  author={P{\'e}rez-Mon, Olaya and del Coz, Juan Jos{\'e} and Gonz{\'a}lez, Pablo},
  journal={arXiv preprint arXiv:2501.13638},
  year={2025}
}

@misc{uci_emg_dataset,
  title        = {{EMG Data for Gestures}},
  author       = {{UCI Machine Learning Repository}},
  year         = {2022},
  howpublished = {\url{https://archive.ics.uci.edu/dataset/481/emg+data+for+gestures}},
  note         = {Accessed: 2024-05-01}
}

@inproceedings{esuli2018_rnnquant,
  title={A recurrent neural network for sentiment quantification},
  author={Esuli, Andrea and Moreo Fern{\'a}ndez, Alejandro and Sebastiani, Fabrizio},
  booktitle={Proceedings of the 27th ACM international conference on information and knowledge management},
  pages={1775--1778},
  year={2018}
}

@misc{human_activity_recognition_using_smartphones_240,
  author       = {Reyes-Ortiz, Jorge and Anguita, Davide and Ghio, Alessandro and Oneto, Luca and Parra, Xavier},
  title        = {Human Activity Recognition Using Smartphones},
  year         = {2013},
  howpublished = {UCI Machine Learning Repository},
  note         = {DOI: \url{https://doi.org/10.24432/C54S4K}}
}

@inproceedings{esuli2022concise,
  title={A concise overview of LeQua@ CLEF 2022: learning to quantify},
  author={Esuli, Andrea and Moreo, Alejandro and Sebastiani, Fabrizio and Sperduti, Gianluca},
  booktitle={International Conference of the Cross-Language Evaluation Forum for European Languages},
  pages={362--381},
  year={2022},
  organization={Springer}
}

@article{gonzalez2024binary,
  title={Binary quantification and dataset shift: an experimental investigation},
  author={Gonz{\'a}lez, Pablo and Moreo, Alejandro and Sebastiani, Fabrizio},
  journal={Data Mining and Knowledge Discovery},
  volume={38},
  number={4},
  pages={1670--1712},
  year={2024},
  publisher={Springer}
}

@article{saerens2002adjusting,
  title={Adjusting the outputs of a classifier to new a priori probabilities: a simple procedure},
  author={Saerens, Marco and Latinne, Patrice and Decaestecker, Christine},
  journal={Neural computation},
  volume={14},
  number={1},
  pages={21--41},
  year={2002},
  publisher={MIT Press}
}

@article{vaswani2017attention,
  title={Attention is all you need},
  author={Vaswani, Ashish and Shazeer, Noam and Parmar, Niki and Uszkoreit, Jakob and Jones, Llion and Gomez, Aidan N and Kaiser, {\L}ukasz and Polosukhin, Illia},
  journal={Advances in neural information processing systems},
  volume={30},
  year={2017}
}

@article{lim2021_tft,
  title={Temporal fusion transformers for interpretable multi-horizon time series forecasting},
  author={Lim, Bryan and Ar{\i}k, Sercan {\"O} and Loeff, Nicolas and Pfister, Tomas},
  journal={International journal of forecasting},
  volume={37},
  number={4},
  pages={1748--1764},
  year={2021},
  publisher={Elsevier}
}

@article{sebastiani2020evaluation,
  title={Evaluation measures for quantification: An axiomatic approach},
  author={Sebastiani, Fabrizio},
  journal={Information Retrieval Journal},
  volume={23},
  number={3},
  pages={255--288},
  year={2020},
  publisher={Springer}
}

@article{kullback1951information,
  title={On information and sufficiency},
  author={Kullback, Solomon and Leibler, Richard A},
  journal={The annals of mathematical statistics},
  volume={22},
  number={1},
  pages={79--86},
  year={1951},
  publisher={JSTOR}
}

@article{nie2022time,
  title={A time series is worth 64 words: Long-term forecasting with transformers},
  author={Nie, Yuqi and Nguyen, Nam H and Sinthong, Phanwadee and Kalagnanam, Jayant},
  journal={arXiv preprint arXiv:2211.14730},
  year={2022}
}

@inproceedings{zhou2021informer,
  title={Informer: Beyond efficient transformer for long sequence time-series forecasting},
  author={Zhou, Haoyi and Zhang, Shanghang and Peng, Jieqi and Zhang, Shuai and Li, Jianxin and Xiong, Hui and Zhang, Wancai},
  booktitle={Proceedings of the AAAI conference on artificial intelligence},
  volume={35},
  number={12},
  pages={11106--11115},
  year={2021}
}

@inproceedings{lin2017focal,
  title={Focal loss for dense object detection},
  author={Lin, Tsung-Yi and Goyal, Priya and Girshick, Ross and He, Kaiming and Doll{\'a}r, Piotr},
  booktitle={Proceedings of the IEEE international conference on computer vision},
  pages={2980--2988},
  year={2017}
}

@misc{smartfallmm_dataset,
  title        = {SmartFallMM Dataset},
  howpublished = {\url{https://github.com/txst-cs-smartfall/SmartFallMM-Dataset}},
  note         = {Accessed: 2026-02-28},
  author       = {{Texas State University SmartFall Lab}},
  year         = {2026}
}

@inproceedings{kabir2025transconv,
  title={TransConv-DDPM: Enhanced Diffusion Model for Generating Time-Series Data in Healthcare},
  author={Kabir, Md Shahriar and Alamgeer, Sana and Debnath, Minakshi and Ngu, Anne HH},
  booktitle={2025 IEEE 49th Annual Computers, Software, and Applications Conference (COMPSAC)},
  pages={866--875},
  year={2025},
  organization={IEEE}
}

@inproceedings{mithila2025ms,
  title     = {{MS-SCANet}: A multiscale transformer-based architecture with dual attention for no-reference image quality assessment},
  author    = {Mithila, Mayesha Maliha R. and Farias, Myl{\`e}ne C.Q.},
  booktitle = {Proc. IEEE Int. Conf. on Acoustics, Speech and Signal Processing ({ICASSP})},
  year      = {2025},
  pages     = {1--5},
  publisher = {IEEE}
}

\end{document}